\documentclass[sn-mathphys-num]{sn-jnl}

\usepackage{graphicx}%
\usepackage{multirow}%
\usepackage{amsmath,amssymb,amsfonts}%
\usepackage{amsthm}%
\usepackage{mathrsfs}%
\usepackage[title]{appendix}%
\usepackage{xcolor}%
\usepackage{textcomp}%
\usepackage{manyfoot}%
\usepackage{booktabs}%
\usepackage{algorithm}%
\usepackage{algorithmicx}%
\usepackage{algpseudocode}%
\usepackage{listings}%
\usepackage{lmodern}
\usepackage{fix-cm}

\begin{document}

\title[Article Title]{Cross-Subject Domain Adaptation for Classifying Working Memory Load with Multi-Frame EEG Images}


\author[1]{\fnm{Junfu} \sur{Chen}}\email{cjf@nuaa.edu.cn}

\author[2]{\fnm{Sirui} \sur{Li}}\email{072440202@nuaa.edu.cn}
\author*[1]{\fnm{Dechang} \sur{Pi}}\email{pinuaa@nuaa.edu.cn}



\affil[1]{\orgdiv{College of Computer Science and Technology}, \orgname{Nanjing University of Aeronautics and Astronautics},  \city{Nanjing}, \postcode{211106}, \country{China}}
\affil[2]{\orgdiv{College of Civil Aviation}, \orgname{Nanjing University of Aeronautics and Astronautics},  \city{Nanjing}, \postcode{211106}, \country{China}}

\abstract{Working memory (WM), denoting the information temporally stored in the mind, is a fundamental research topic in the field of human cognition. Electroencephalograph (EEG), which can monitor the electrical activity of the brain, has been widely used in measuring the level of WM. However, one of the critical challenges is that individual differences may cause ineffective results, especially when the established model meets an unfamiliar subject. In this work, we propose a cross-subject deep adaptation model with spatial attention (CS-DASA) to generalize the workload classifications across subjects. First, we transform EEG time series into multi-frame EEG images incorporating spatial, spectral, and temporal information. First, the Subject-Shared module in CS-DASA receives multi-frame EEG image data from both source and target subjects and learns the common feature representations. Then, in the subject-specific module, the maximum mean discrepancy is implemented to measure the domain distribution divergence in a reproducing kernel Hilbert space, which can add an effective penalty loss for domain adaptation. Additionally, the subject-to-subject spatial attention mechanism is employed to focus on the discriminative spatial features from the target image data. Experiments conducted on a public WM EEG dataset containing 13 subjects show that the proposed model is capable of achieving better performance than existing state-of-the-art methods. 
}




\maketitle

\section{Introduction}\label{sec1}

Electroencephalogram (EEG) is a physiological electrical signal with corresponding characteristics generated by the brain receiving certain stimuli. It not only can effectively reflect the functional state of the brain, but also give feedback on the current state of a person's physical function \cite{I1}, and is therefore widely used in the analysis of neurological diseases \cite{I2}, brain-computer interfaces \cite{I3}, and the study of cognitive processes \cite{I4}. With the help of EEG devices, it is convenient to obtain event-related potential data (ERP) in visual working memory tasks. It has been found that there is a strong correlation between individual working memory capacity and the signals produced by the brain nervous system that maintains memory over species. EEG-based class  ification methods \cite{I5,I6,add1} are important for further analysis of the mental activity during working memory.

Traditional methods for detecting the mental workload in WM tasks always focus on extracting hand-crafted features and using machine learning algorithms. For example, alpha event-related desynchronization has been one of the most popular features that show a high correlation with WM performance \cite{I7}. Additionally, it was also found that the posterior alpha oscillations before the memory phase influence the prediction accuracy \cite{I8}. In view of time-frequency characteristics, wavelet entropy has shown its effectiveness in evaluating mental workload in WM tasks \cite{I9}. Another one of the most commonly used features in assessing mental workload is the power spectral density (PSD)\cite{I10,b32}. On the basis of PSD, some studies \cite{I10, I11} explored applying the Shannon entropy to measuring the uniformity of the PSD. This method, namely spectral entropy, can capture the difference between the EEG power spectrum distribution groups. And hence, the spectral entropy is qualified for a classification standard when predicting individual performance in many scenarios including sensorimotor rhythm-based BCI \cite{I12}, WM tasks et al. In some WM tasks with high-dimensionality systems, the Lempel-Ziv complexity \cite{I13, I14} has the capability of improving the classification accuracy and has the advantage on easier implementation for real-time use. 

Recently, as the demand for signal analysis accuracy increases and computer computing power develops \cite{chen2024multi}, deep learning has become the dominant tool for EEG signal analysis. Hsu \cite{I15} presented a fuzzy neural network based on multi-layer perceptron architecture and and the results show its higher accuracy than traditional neural networks on the task of single-trail motor imagery EEG classification. To observe spatial features from EEG signals, Tang et al. \cite{I16} proposed a deep convolution neural network (CNN) and demonstrated better performance than the traditional support vector machine-based methods. To enhance temporal information, Zhang et al. \cite{I17} combined convolution networks and recurrent neural networks in a cascade or parallel manner.

These data-driven approaches have been extensively studied in subject-dependent scenarios, and experimental results show that they can perform well for specific subjects for tasks such as EEG signal feature extraction and classification. However, in practice, potential feature discrepancies between subjects lead to unsatisfactory results for subsequently introduced individuals (subjects) on previously trained models. One burdensome solution is to calibrate the model with a lot of labeled data on these subjects, which is a time-consuming and expensive mission. Another possible solution is to utilize transfer learning methods to explore the possibility of matching the distributions between source and target subjects. Nonetheless, a majority of transfer learning methods for EEG lack decent ability to understand multi-source features with spatial, temporal, and spectral information, and how to use transfer learning to solve brain state recognition tasks in the working memory EEG field is rarely investigated. 

Inspired by recent studies on the transferability between subjects in EEG classification tasks, in this paper, we propose a cross-subject deep adaptation network with a spatial attention strategy named CS-DASA that can effectively transfer task models between subjects. The key idea of this work is to establish a neural network architecture that can effectively extract robust features and accomplish subject-to-subject transfer learning. To achieve this goal, first, we transform the original working memory EEG time series into multi-frame EEG spectral images, which can incorporate spectral, spatial, and temporal information. Then, CS-DASA inputs the EEG image pairs from both subjects together into several convolutional long short-term memory (ConvLSTM) layers, which extract the shared feature representation by freezing the parameters. Subsequently, the learned feature representations are input into subject-specific convolution layers, in which the parameters can be trained according to different subjects. Meanwhile, the domain discrepancy from this pair of subjects can be calculated from each couple of subject-specific layers by multi-kernel maximum mean discrepancies (MK-MMD) in a reproducing kernel Hilbert space \cite{b18}. And then, a joint optimization incorporating the goal of reducing the domain discrepancy and task loss of the source domain is conducted for the cross-subject domain adaptation. Additionally, an attention mechanism with the capacity of focusing on the discriminative spatial information between two subject domains is implemented to improve the adaptation performance. The main contributions of this paper are summarized as follows:

\begin{itemize}

\item As far as we know, few works have investigated the model transferability for the EEG-based working memory load tasks. We conduct comprehensive experiments on an open-access data set with the cross-subject transfer learning setting, which is rarely studied by other researchers. 

\item EEG signals do show advantage on owning high temporal resolution, but they are observed with low spatial resolution. To enhance the spatial information, we apply a topology-preserved projection method to converting 3-D original electrode positions into points on 2-D space and employ an interpolation technique to form EEG images that contain multi-source information from the perspective of time, frequency, and space.

\item The proposed CS-DASA can capture the spatial similarity between the source and target subjects with a global spatial attention mechanism. Incorporating the relevant features from the target subjects can improve the classification performance for the transfer learning since the relevant features utilized in the training phase can help the model understand the target subjects when evaluating the WM load in the testing phase. 
\end{itemize}

The structure of this paper is organized as follows: Section \uppercase\expandafter{\romannumeral2} discusses the related works; Section \uppercase\expandafter{\romannumeral3} briefly introduces the task definitions and details the proposed transfer learning framework; Section \uppercase\expandafter{\romannumeral4} gives the dataset and the experimental results; Section \uppercase\expandafter{\romannumeral5} discusses the results and presents possible future directions; Section \uppercase\expandafter{\romannumeral6} shows the conclusion.

\section{Related Work}\label{sec2}
\subsection{Deep learning for EEG working memory load classification}
Working memory, denoting the information temporally stored in the brain, is a fundamental research topic for human cognition. With the characteristics of relatively low expense for collecting and high temporal resolution, EEG has been widely used in working memory tasks, especially in the classification of mental workload. As mentioned in the paragraphs, in need of better model performance and an end-to-end structure, some deep learning-based studies have been proposed to solve WM load classification tasks. Jiao et al. \cite{R1} proposed to introduce a gated Boltzmann machines component to enhance CNN models. The results show their models can even achieve better performance than deep recurrent CNNs. Kuanar et al.\cite{R2} explored the performance of a hybrid recurrent neural network for predicting the levels of cognitive load. Yang et al. \cite{R3} introduced a feature mapping layer and utilized an ensemble autoencoders to preserve the local information from EEG recordings. The results show the proposed model can perform better than some classical MW estimators. Zhang et al. \cite{R4} designed a deep recurrent 3D convolutional neural network architecture to simultaneously capture the spatial, spectral, and temporal dimensions, which shows higher accuracy for cross-task conditions of mental workload identification. The diver’s cognitive workload under different workload tasks has also been studied with CNN models in some works \cite{R5, R6}. Che et al. \cite{b33} proposed to use graph convolutional layers to extract discriminative features, which shows better performance in color decoding for different visual WM stages.

\subsection{Transfer learning for EEG analysis }
A large number of traditional methods, such as Kernel common spatial patterns (CSP) \cite{R7,R8}, Riemannian space \cite{R9,R10}, and PCA-based methods \cite{R11,R13}, have been proposed to explore the possibility of matching the distributions between source and target subjects. Dai et al. \cite{R7} proposed a CSP-based method to conduct transfer learning on the dataset IVa of BCI Competition. This method employs a domain-invariant kernel trick to learn the distribution difference between subjects. 
Zanni et al.\cite{R9} introduced Riemannian Gaussian distributions to define the covariance matrices. With the help of the covariance matrices in the Riemannian manifold, data between different sessions/subjects become comparable.
Jiang et al. \cite{R11} constructed a latent subspace using transfer component analysis (TCA) on an open-access scalp EEG dataset. The results show that the built subspace with TCA can make the model perform better than the original feature space. Li et al. \cite{R14} designed a multi-source transfer learning method, integrating the source models to calibrate different sessions. In the fast-deployment scenarios, the proposed method has a significant advantage over other methods. 

Meanwhile, deep learning, as a class of end-to-end methods with powerful feature extraction capability, has also been gradually applied to EEG transfer learning. Yin and Zhang \cite{R15} presented an adaptive Stacked Denoising AutoEncoder (SDAE) for cross-session tasks on Mental Workload EEG signals. Experiments show that SDAE can also achieve success for online implementation. Li et al. \cite{R16} used neural networks to conduct EEG latent representations and optimized the model through improving the similarity between source and target data representation in the way of adversarial training. Inspired by the fact that the left and right hemispheres of the brain are asymmetric to emotional response, Li et al. \cite{R17} proposed a domain adversarial neural network to explore the domain difference from each hemisphere, which can extract robust features for a different subject. Following the idea of the generative adversarial network, subject adaptation network \cite{R18} was proposed to mitigate cross-subject EEG variance. This model considers a generative adversarial network from a wider perspective and shows effectiveness on a VEP oddball task.

\section{methodology}\label{sec3}
\subsection{Task Paradigm and Related Definitions}
\begin{figure}[htbp]
\centerline{\includegraphics[width=0.9\textwidth]{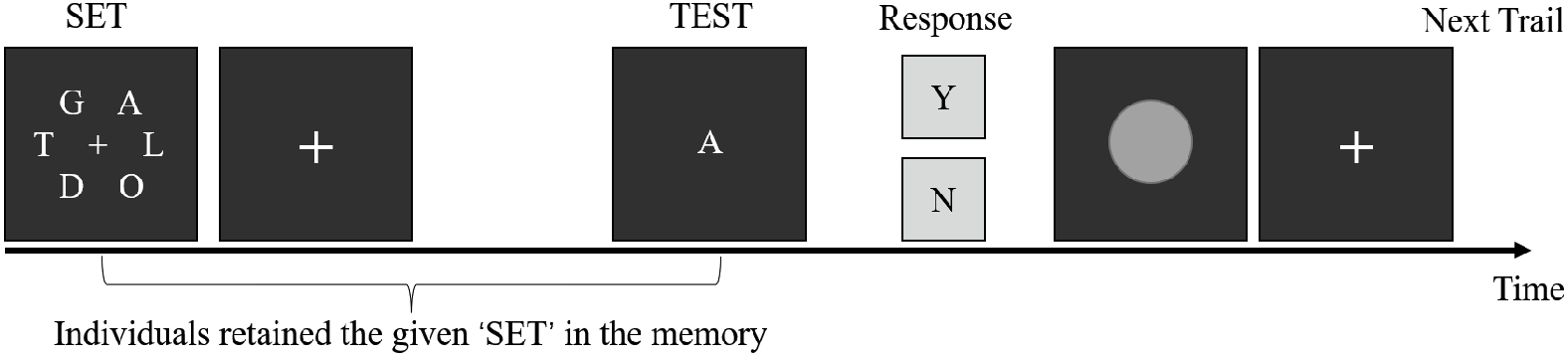}}
\caption{Time course of a visual working memory task}
\label{finalfig1}
\end{figure}

Before introducing the details of methodology, we first briefly describe the experiment procedure of a prototype of WM tasks and give some research-related definitions. Fig.~\ref{finalfig1} shows the time course of a widely-used visual WM task paradigm designed by Bashivan et al. \cite{I5} In the experiment, a ‘SET’ usually includes several English characters and appears around a center point for a brief period in the beginning of each trail. Then, the participates can only see a cross in the middle of the screen for a longer period. Next, a ‘TEST’ character appears on the screen for the participates to judge whether this character existed in the ‘SET’ through pressing a button (‘Y’ or ‘N’). At last, the corresponding feedback with a form of green or red circle shows on the screen and a cross follows this step. The size of a ‘SET’ for each trail should be chosen at random and only the correct response trails will be considered in the further study.

For a task of WM load level classification, the duration of what researchers focus on is the period (T-second long) for the participates to remember the information in their memory. And hence, each channel of an EEG trail owns a recording time sequence including $T\times f$ time points, where $f$ denotes the sampling rate of the BCI system. However, instead of analyzing raw time sequences, researchers show more interest in studying the spectral information from EEG signals for WM tasks. In most WM tasks, three kinds of frequency bands including theta (4-7Hz), alpha (8-13Hz) and beta (13-30 Hz) are mainly measured for further research. The following are several task-related definitions.

Definition 1 (EEG classification with domain transfer): Given a completely-labeled source domain $\mathcal{D}_s=\left\{ \left( X_s^i,y_s^i \right) \right\} _{i=1}^{N}$ and a target domain $\mathbb{D}_t$ including $N_l$ ($N_l$ can be 0) samples with labels $\left\{ \left( X_t^i,y_t^i \right)\right\}_{i=1}^{N_l}$ and $N_u$ samples without any labels $\left\{ X_t^i \right\}_{i=N_l+1}^{N_l+N_u}$, EEG classification transfer learning hopes to utilize the learned knowledge $f:X_s\mapsto y_s $ to acquire the mapping function $f:X_t\mapsto y_t$ in the target domain. Additionally, the promise is that $\mathcal{X}_s\ne{X_s} $, ${y_t}\ne{y_s} $, $P_s\left(X\right)\ne P_t\left(X\right) $, and/or $P_s\left( y|X  \right)\ne P_t\left( y|X \right) $, where $\mathcal{X}$ and $\mathcal{Y}$ represent the feature spaces of $X$ and $y$, $P\left( X \right) $ means the marginal probability distribution, and $P\left( y|X \right)$ refers to the conditional probability distribution.

Definition 2 (One-to-One transfer): In this scenario, the target is one subject’s EEG, and the source is another subject’s EEG. One-to-One Transfer makes sense when only one existing subject is available. In the following sections, we denote the One-to-One Transfer as $\mathcal{O}\rightarrow  \mathcal{O}$.

Definition 3 (Many-to-One transfer): The target is one subject’s EEG, and the source is another several subjects’ EEG. Many-to-One Transfer makes sense when multiple existing subjects are available. In the following sections, we denote the Many-to-One Transfer as $\mathcal{M}\rightarrow \mathcal{O}$.
\subsection{Pipeline Overview}
\begin{figure*}[htbp]

\centerline{\includegraphics[width=0.95\textwidth]{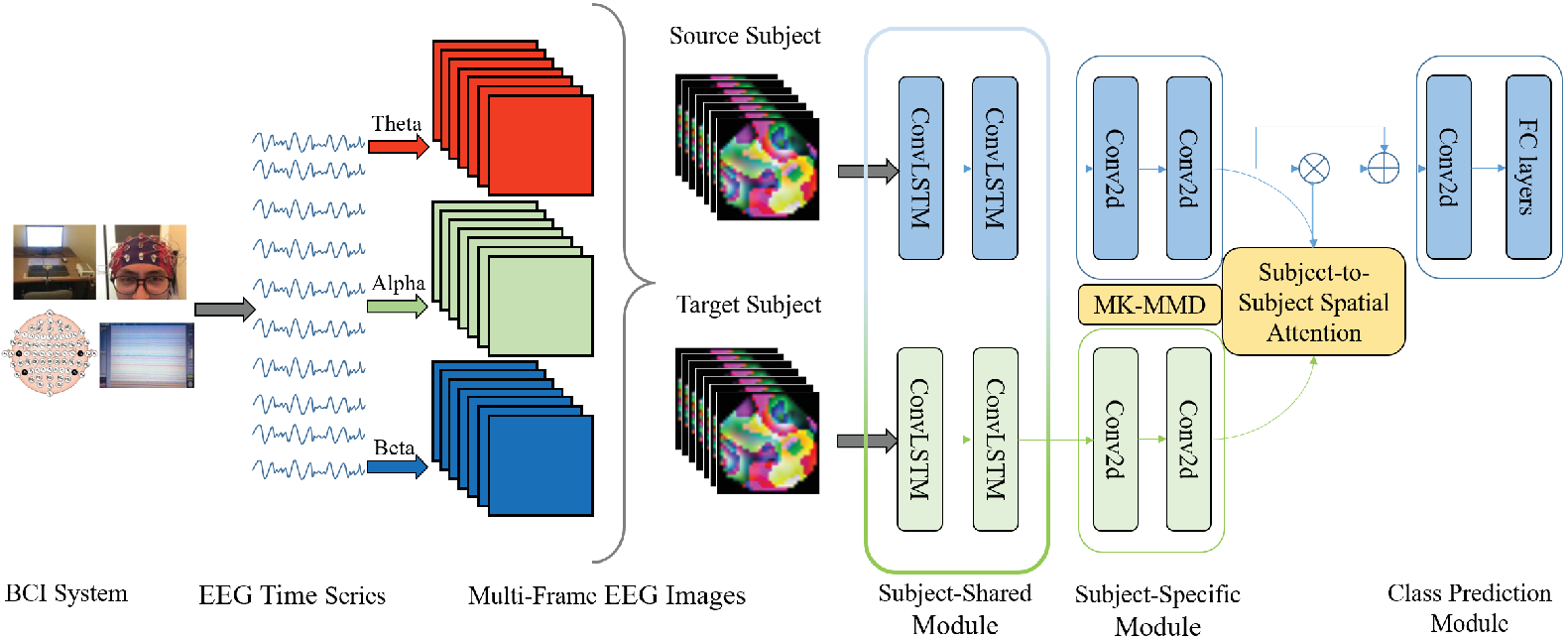}}
\caption{Pipeline overview of the proposed cross-subject transfer model: (1) EEG time series are divided into 7 slices of 0.5-second windows and spectral power within the three frequency bands (theta (4-7Hz), alpha (8-13Hz), and beta (13-30 Hz)) are extracted by Fast Fourier Transform (FFT);
(2)	A 2-D space position map of EEG electrodes formed by the azimuthal equidistant projection is then utilized to make 7-frame EEG images with three spectral channels;
(3)	Then, a pair of source and target EEG images are fed into the Subject-Shared module. This Subject-Shared module has been pretrained only by the source data and is frozen in the training phase;
(4)	Subsequently, the output feature pair is input in the Subject-Specific module. In this module, the features from different domains are further fed into the same layer architecture but with different network parameters; 
(5)	Finally, the Class-Prediction module processes the output features by the last module and makes a decision for the level of WM load. 
 }
\label{finalfig2}
\end{figure*}

The overall framework of the proposed CS-DASA is shown in Fig.~\ref{finalfig2}. As shown in the figure, the CS-DASA mainly consists of three parts: a shared feature representation module with several ConvLSTM layers, subject-specific feature extraction module through several 2-D convolution layers and a class prediction module.

\subsection{Making Multi-frame EEG Images}
Compared with other brain imaging methods, EEG has higher temporal resolution, so most of the proposed research methods are based on temporal domain and spectral domain to analyze the EEG signals. In order to allow the model to observe more spatial information and extract multi-source features consisting of spatial domain, temporal domain and spectral domain, we transform the continuous time window EEG signals into multi-channel multi-frame EEG images, where the channels represent the three frequency bands (theta (4-7Hz), alpha (8-13Hz) and beta (13-30 Hz)) mentioned above, and the frames represent time slices in each trial.

To obtain more spatial features with topology information from EEG signals, inspired by \cite{b28}, we project electrode coordinates in 3-D space onto the 2-D plane using a projection method called Azimuthal Equidistant projection derived from cartography.

Azimuthal Equidistant projection \cite{b29} is one form of the spherical projection, and hence has the azimuthal characteristic, which means that all directions can be preserved from the perspective of the center point of the projection. In terms of this characteristic, the distance measured between the given center and other points are at true scale. For better illustrating how an EEG electrode will be projected, we give a simplified figure shown in Fig. \ref{sphere}, where all EEG electrode nodes exist on a sphere.  Then, the position after the azimuthal equidistant projection can be calculated:

\begin{figure}[htbp]

\centerline{\includegraphics[width=0.9\textwidth]{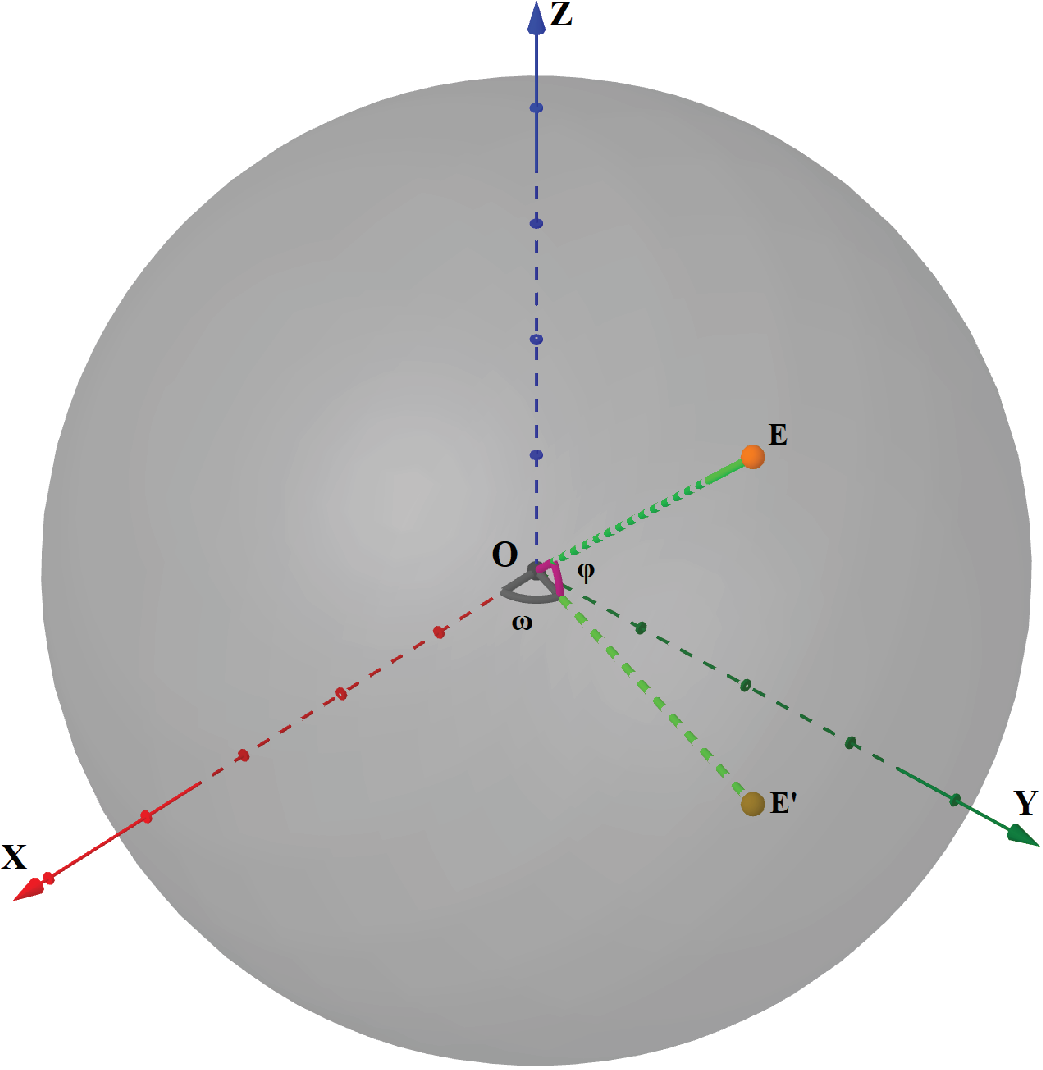}}
\caption{A simplified example for an EEG node with Azimuthal Equidistant projection. $E$, holding the coordinate $\left(x,y,z\right)$, represents an electrode node from the BCI system. $E^{'}$ is the projection of $E$ on the plane $XOY$ and its coordinate is $\left(x,y,0\right)$.  }
\label{sphere}
\end{figure}

\begin{equation}
\left\{ \begin{array}{l}
	\varphi =\arctan \frac{y}{x}\\
	\omega =\arctan \frac{z}{\sqrt{x^2+y^2}}\\
	\rho =r\times \left( \frac{\pi}{2}-w \right)\\
	x_p=\rho \cos \varphi\\
	y_p=\rho \sin \varphi\\
\end{array} \right. 
\end{equation}

\noindent where $r$ denotes the radius, and $(x_p,y_p)$ represents the projected position of point $E$ through Azimuthal Equidistant projection. 

\begin{figure*}[htbp]

\centerline{\includegraphics[width=0.95\textwidth]{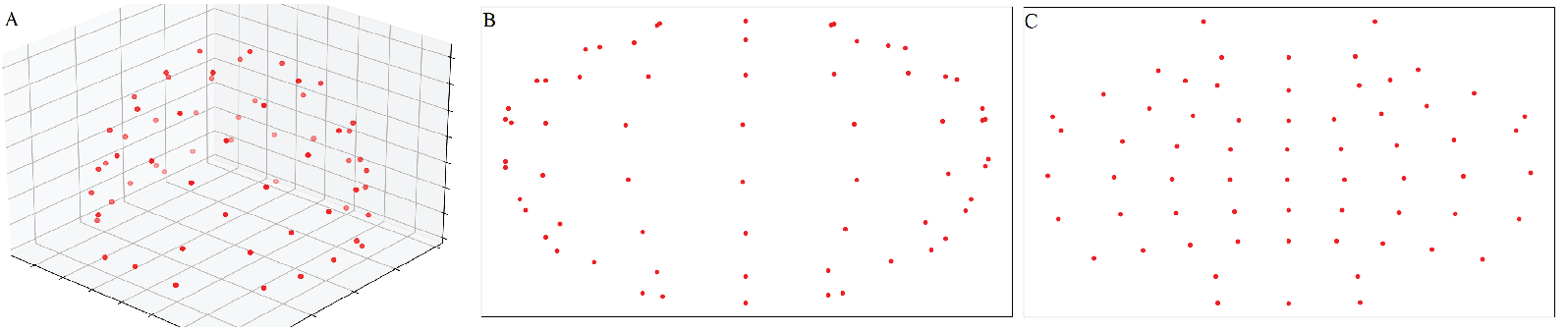}}
\caption{Different position presentations for EEG electrode nodes}
\label{positionfig}
\end{figure*}

Fig.~\ref{positionfig} demonstrates three different space presentations of EEG nodes, including the true positions of electrodes from a BCI system (Fig.~\ref{positionfig}.A), the direct projection on the $XOY$ plane without preserving topology relationships (Fig.~\ref{positionfig}.B), and the Azimuthal Equidistant projection with topology information between nodes (Fig.~\ref{positionfig}.C). Although the projected EEG nodes have owned the spatial topology information, the features they can contain are too sparse for image classification techniques. To solve this problem, we utilize the CloughTocher scheme \cite{b30} to interpolate the spectral domain signals from each EEG nodes, and then an information map with the size of $32\time 32$ can be generated. For each frequency band, the same procedure is conducted, which finally can form an image with three channels.

\subsection{Subject-Shared Representation Learning with ConvLSTM}
The first part of the proposed model is composed of several ConvLSTM layers aimed at extracting common representation features from both subjects. Note that before subject-subject transfer these ConvLSTM layers have been trained in the source domain, while their parameters will be frozen during the transfer learning. ConvLSTM implements convolution operations for input-to-state and state-to-state transitions \cite{b20}, which can capture more spatial information, like topology structure between electrode locations, of multi-frame EEG images than LSTM and extract richer temporal information than convolution neural networks.

ConvLSTM replaces matrix multiplication with a convolution operation for each gate in the LSTM cell. In this way, it captures the underlying spatial features by performing convolution operations in multidimensional data. Another major difference between ConvLSTM and LSTM is the number of input dimensions. Unlike most LSTMs that process sequential input like time series and text embeddings, the ConvLSTM layer used in this paper can accept video-like tensors consisting of multi-frame EEG images with three channels in each time step.
One whole step of a ConvLSTM layer can be expressed as follows.
\begin{equation}
\begin{split}
&\boldsymbol{i}_t=\sigma \left(\boldsymbol{W}_{ix}*\boldsymbol{x}_t+\boldsymbol{W}_{ih}*\boldsymbol{h}_{t-1}+\boldsymbol{W}_{ic}\circ \boldsymbol{c}_{t-1}+\boldsymbol{b}_i \right)\\
&\boldsymbol{f}_t=\sigma \left( \boldsymbol{W}_{fx}*\boldsymbol{x}_t+\boldsymbol{W}_{fh}*\boldsymbol{h}_{t-1}+\boldsymbol{W}_{fc}\circ \boldsymbol{c}_{t-1}+\boldsymbol{b}_f \right)
\\
&\boldsymbol{o}_t=\sigma \left( \boldsymbol{W}_{ox}*x_t+\boldsymbol{W}_{oh}*\boldsymbol{h}_{t-1}+\boldsymbol{W}_{oc}\circ \boldsymbol{c}_{t-1}+\boldsymbol{b}_o \right) 
\\
&g_t=\tanh \left( \boldsymbol{W}_{gx}*\boldsymbol{x}_t+\boldsymbol{W}_{gh}*\boldsymbol{h}_{t-1}+\boldsymbol{b}_g \right) 
\\
&\boldsymbol{c}_t=f_t \circ c_{t-1} + i_t\circ g_t
\\
&\boldsymbol{h}_t=\boldsymbol{o}_t\circ \tanh \left( \boldsymbol{c}_t \right) 
\end{split}
\end{equation}
\noindent where $t$ denotes the $t$th step of ConvLSTM; $ \boldsymbol{x}_t$ denotes the input data; $ \boldsymbol{h}_t$ denotes the hidden state; $ \boldsymbol{c}_t$ denotes the state of the storage cell; $ \boldsymbol{i}_t$, $ \boldsymbol{f}_t$ and $ \boldsymbol{o}_t$ are the input gate, forget gate and output gate of ConvLSTM, respectively. $ \boldsymbol{W}_t$ and $ \boldsymbol{b}_t$ are the weights and biases to be learned; $*$, $\circ$, $\sigma$ and tanh denote the convolution operation, element multiplication, Sigmoid function and tanh function. Let mapping function $f_{cl}^i \left( \cdot \right) $ denote the $i$th layer of the $N$ stacked ConvLSTM layers, and then the representation features of source and target subject EEG image data can be formulated as:
\begin{equation}
\begin{split}
\boldsymbol{F}_{S}^{cl}=f_{cl}^{N}\left( f_{cl}^{N-1}\left( ...f_{cl}^1\left( \mathcal{X}_S \right) \right) \right)\\
\boldsymbol{F}_{T}^{cl}=f_{cl}^{N}\left( f_{cl}^{N-1}\left( ...f_{cl}^1\left( \mathcal{X}_T \right) \right) \right)
\end{split}
\end{equation}
\noindent where $\boldsymbol{F}_{S}^{cl}$ and $\boldsymbol{F}_{T}^{cl}$ are final feature representations of the two subject domain through $N$ ConvLSTM layers.

\subsection{Subject-Specific Knowledge Transfer with MK-MMD}
The output $\boldsymbol{F}_{S}^{cl}$ and $\boldsymbol{F}_T^{cl}$ through stacked ConvLSTM layers are then input their subject-specific feature extraction module consists of several Conv2D layers. MK-MMD strategy \cite{b18} will impose constraints for subject-specific feature extraction during transfer learning. Through embedding the learned representations output by subject-specific layers from two subject domains to a reproducing kernel Hilbert space, MK-MMD can reduce the domain discrepancy with the help of adaptation network described in Fig.~\ref{finalfig1}. And, the squared formulation of MK-MMD can be calculated as:
\begin{equation}
d_{MMD}^{2}\left( \mathcal{D}^S,\mathcal{D}^T \right) =\lVert \frac{1}{n}\sum_{i=1}^n{\phi \left( e_{i}^{S} \right) -\frac{1}{m}\sum_m^1{\phi \left( e_{i}^{T} \right)}} \rVert _{\mathcal{H}}^{2}
\end{equation}
\noindent where $\mathcal{H}$ denotes the reproducing kernel Hilbert space and $\phi  \left(\cdot \right)$ represents the kernel function endowed with a Gaussian kernel in this work. $e_{i}^{S}$ and $e_{i}^{T}$ denote the feature samples from two subject latent representation set, $\mathcal{D}^S$ and $\mathcal{D}^T$.
Since the output from the previous layers is 4-D data with the size of $t \times c \times w \times h$, we reshape them into $(t*c)\times w \times h$ in order to make their size suitable for Conv2D layers. Let $f_{2d}^{i}\left( \cdot \right)$ denote the $i$th layer of Conv2D, then the transfer loss $l_{MMD}$ from MK-MMD can be calculated as:
\begin{equation}
\begin{split}
\mathcal{D}_{i}^{S}=f_{2d}^{i}\left( f_{2d}^{i-1}\left( ...f_{2d}^{1}\left( \boldsymbol{F}_{S}^{cl} \right) \right) \right)\\
\mathcal{D}_{i}^{T}=f_{2d}^{i}\left( f_{2d}^{i-1}\left( ...f_{2d}^{1}\left( \boldsymbol{F}_{T}^{cl} \right) \right) \right)\\
l_{MMD}=\sum_{i=1}^N{d_{MMD}^{2} \left( \mathcal{D}_{i}^{S} , \mathcal{D}_{i}^{T} \right)}
\end{split}
\end{equation}

It is worth noting that unlike in the Subject-Shared module where the source and target subjects share the same parameters of $f_{cl}^i \left( \cdot \right) $, the parameters are independent in $f_{2d}^i\left( \cdot \right) $ and we give the same formulation for simplicity.
\subsection{Subject-to-Subject Spatial Attention}
During visual working memory tasks, each subject wears the BCI devices in the uniform norms. And hence, the time series EEG signals observed from each electrode share similar patterns since these electrodes monitor the fixed parts of the brain. EEG images made by the proposed method can preserve the original spatial information like spatial correlations and topology relationships to a great degree. It is worthwhile to explore and utilize the feature correlation between the source and target subject to enhance the performance of transfer learning. Therefore, we propose the subject-to-subject spatial attention to allocate importance from each pair of regions of the source and target subjects. Specifically, we design the model architecture to make the size  of feature matrix ($w\times h $) the same with the raw input EEG image, which may let the latent representation align the spatial structure of the raw image as much as possible.  Before calculating the attention matrices, it is necessary to reshape the output features from Conv2D layers:
\begin{equation} 
\begin{split}
&\boldsymbol{F}_{S}^{2d}=f_{2d}^{N}\left( f_{2d}^{N-1}\left( ...f_{2d}^1\left( \mathcal{X}_S \right) \right) \right)\\
&\boldsymbol{F}_{T}^{2d}=f_{2d}^{N}\left( f_{2d}^{N-1}\left( ...f_{2d}^1\left( \mathcal{X}_T \right) \right) \right)\\
\end{split}
\end{equation}

\begin{equation}
\begin{split}
&\boldsymbol{F}_{S}^{2d}:\mathbb{R}^{c\times w\times h}\xrightarrow[]{} \mathbb{R}^{c\times L}\\
&\boldsymbol{F}_{T}^{2d}:\mathbb{R}^{c\times w\times h}\xrightarrow[]{}\mathbb{R}^{c\times L}\\
&\mathcal{A}=Softmax \left( \boldsymbol{F}_{S}^{{2d}^{'}} \otimes \boldsymbol{F}_{T}^{2d} \right)\\
&\boldsymbol{F}_{S}^{att} = \boldsymbol{F}_{S}^{2d} \otimes \mathcal{A}\\
&\boldsymbol{F}_{S}^{att}:\mathbb{R}^{c\times L}\xrightarrow[]{}\mathbb{R}^{c\times w\times h}\\
&\boldsymbol{F}_{S}^{o}=\boldsymbol{F}_{S}^{2d}\oplus \boldsymbol{F}_{S}^{att} 
\end{split}
\end{equation}
\noindent where $c$, $w$ and $h$ denote channel, width and height of the output feature from Conv2D; $N$ is equal to $w\times h$; $\mathcal{A}$ represents the calculated attention matrix.  $\boldsymbol{F}_{S}^{{2d}^{'}}$ represents the matrix transposition of $\boldsymbol{F}_{S}^{2d}$ and $\otimes$ is the dot-product operation. And then, the final feature representation $\boldsymbol{F}_{S}^{o}$ incorporates the subject-to-subject spatial information $\boldsymbol{F}_{S}^{att}$. Note that the concatenation operation between $\boldsymbol{F}_{S}^{2d}$ and $\boldsymbol{F}_{S}^{att} $ is in the dimension $c$ of $c\times w\times h$, and the size of $\boldsymbol{F}_{S}^{o}$ is $c\times w\times h$.  
\subsection{Total Loss Function}
The total loss function of CS-DASA can be divided into domain loss and MMD loss:
\begin{equation}
l_{total}=\frac{1}{N_S}\sum_{i=1}^{N_S}{H\left( y_{i}^{S},\tilde{y}_{i}^{S} \right)}+\gamma l_{MMD}
\end{equation}
\noindent where $H$ denotes the cross-entropy loss, $y_{i}^{S}$ is the source domain label, $\tilde{y}_{i}^{S}$ represents the output of the source domain, and $\gamma$ is the domain discrepancy penalty parameter.

\section{Experiments}

\begin{table*}[!ht]
\caption{Cross-Subject $\mathcal{O}\rightarrow \mathcal{O}$ Transfer}
\centering
\resizebox{\linewidth}{!}{
\begin{tabular}{c c c c c c c c c c c c c c c}

    \hline
        Target subject & SorOnly & TCA & W-BDA & JDA & CNN-3D & DDC & Deep-Coral & Novel-JDA & CS-DASA$\triangledown$ & CS-DASA \\ \hline
        S1 & 52.4/24.2 & 54.7/28.7 & 50.5/29.4 & 57.8/28.0 & 61.3/16.9 & 63.8/20.2 & 63.5/19.4 & 64.8/18.7 & 65.5/21.2 & \textbf{67.3/21.4} \\ 
        S2 & 51.4/26/2 & 52.9/27.5 & 54.4/29.6 & 58.3/27.4 & 62.8/19.3 & 66.5/20.3 & 64.2/21.2 & 62.2/20.5 & 67.3/21.4 & \textbf{69.0/21.1} \\ 
        S3 & 50.8/18.1 & 46.3/24.0 & 46.9/25.2 & 52.0/22.1 & 57.9/19.2 & 61.9/16.1 & 57.5/15.8 & 58.8/18.7 & 62.5 16.6 & \textbf{63.1/17.0} \\ 
        S4 & 55.8/16.4 & 51.6/22.8 & 49.9/22.4 & 50.4/23.1 & 67.5/15.9 & 70.2/14.3 & 62.8/17.0 & 67.2/15.8 & 73.3/14.6 & \textbf{74.7/14.2} \\ 
        S5 & 58.5/18.5 & 51.6/21.6 & 50.8/22.6 & 51.4/23.9 & 69.1/17.4 & 70.9/18.7 & 66.6/19.4 & 67.9/18.6 & 75.7/19.5 & \textbf{77.0/20.2} \\ 
        S6 & 55.2/21.6 & 49.7/22.8 & 49.4/21.8 & 51.7/22.5 & 60.9/22.2 & 62.9/ 24.4 & 61.2/22.9 & 60.2/23.0 & 62.8/24.4 & \textbf{64.5/23.3} \\ 
        S7 & 50.8/31.4 & 52.0/28.6 & 52.6/29.0 & 50.2/31.2 & 61.7/26.6 & 62.5/ 26.8 & 62.1/25.9 & 63.0/27.1 & 65.3/26.8 & \textbf{66.5/27.2} \\ 
        S8 & 52.2/29.5 & 53.1/29.1 & 54.1/28.9 & 51.6/31.6 & 59.9/28.6 & 63.5/27.1 & 62.3/26.9 & 62.1/27.6 & 66.0/26.6 & \textbf{68.3/25.5} \\ 
        S9 & 52.3/24.7 & 53.1/26.4 & 53.7/27.0 & 50.7/27.5 & 61.7/25.3 & 64.4/25.3 & 63.8/24.5 & 62.4/25.8 & 65.2/25.5 & \textbf{66.1/25.7} \\ 
        S10 & 51.1/16.0 & 51.8/17.3 & 49.7/17.5 & 49.7/18.9 & 60.7/13.9 & 62.7/13.8 & 60.6/12.7 & 60.7/12.1 & 65.2/11.1 & \textbf{65.3/11.0} \\ 
        S11 & 40.1/22.3 & 40.7/21.4 & 38.8/22.4 & 38.4/24.0 & 54.6/20.5 & 53.6/21.2 & 53.9/21.7 & 50.5/21.7 & 55.5/20.0 & \textbf{55.8/19.9} \\ 
        S12 & 32.4/19.3 & 39.8/27.0 & 37.0/29.3 & 37.0/29.4 & 51.5/25.3 & 50.8/26.2 & 49.3/26.7 & 51.2/25.1 & 50.8/25.8 & \textbf{52.0/25.5} \\ 
        S13 & 17.5/22.9 & 24.1/30.0 & 24.8/29.7 & 24.5/29.2 & \textbf{38.7/28.2} & 37.2/29.2 & 36.6/27.8 & 37.4/28.2 & 37.3/28.8 & 36.9/29.4 \\ \hline
        Average & 47.0/25.5 & 47.8/25.7 & 48.0/26.8 & 47.1/26.3 & 56.5/20.8 & 60.8/23.1 & 58.8/22.6 & 59.1/22.8 & 62.5/23.4 & \textbf{63.6/23.5} \\ \hline

    \label{tab1}
    \end{tabular}
}

\end{table*}

\begin{table}[!ht]
\caption{Cross-Subject $\mathcal{M}\rightarrow \mathcal{O}$ Transfer}
    \centering
    \begin{tabular}{c c c c c}
    \hline
        $N_{\mathcal{M}}$  & 3 & 5 & 7 & 9  \\ \hline
        SorOnly & 62.7/30.2 & 72.9/29.4 & 78.1/29.2 & 83.4/25.5  \\ \hline
        CNN-3D & 69.8/27.8 & 77.5/26.9 & 80.4/27.7 & 85.1/26.2  \\
        DDC & 73.6/25.3 & 80.7/25.6 & 82.7/24.3 & 86.2/24.0  \\ 
        Deep-Coral & 72.1/26.4 & 77.8/25.3 & 81.6/26.2 & 86.0/24.7  \\ 
        Novel-JDA & 71.8/26.4 & 78.7/25.7 & 83.1/23.7 & 87.7/22.9  \\ 
        CS-DASA$\triangledown$ & 74.9/24.4 & 81.4/24.0 & 84.0/24.1 & 87.5/22.4  \\ 
        CS-DASA & \textbf{76.6/23.8} & \textbf{82.2/24.7} & \textbf{85.4/22.9} & \textbf{88.7/23.4}  \\ \hline
    \end{tabular}

\label{tab2}
\end{table}
\subsection{Dataset and Model Implementation}
We conduct experiments on an open dataset for a modified visual working memory task. Different from the Sternberg memory task \cite{b26}, this task divided serval stages including ending, maintenance, and recall states in each trail, which benefits for configuring different memory loads and for studying WM. 
Contributed by Bashivan et al. \cite{b19}, this WM task study included 15 graduate students between 24 and 33 years of age as participants. To make the analysis more accurate and effective, data from two subjects were excluded for too many myogenic artifacts existing in their EEG signals. The EEG data were collected around the scalp at standard 10-10 locations \cite{b27} with 64 sintered Ag/AgCl electrodes and a sampling rate of 500Hz. The size of the ‘SET’ for each trail was randomly chosen to be 2,4,6, or 8. And hence, in our paper, the working memory states with 2, 4 6, or 8 characters are denoted as loads 1-4 respectively. As shown in Fig.~\ref{finalfig1}, the EEG recordings used for recognizing the amount of mental workload appear in the period (3.5 s) when participants retained the given ‘SET’ in the memory. The goal of the WM classification task is to identify the load level (1-4) from these EEG recordings. 2670 trials from a total of 3120 recorded trials are chosen for this cross-subject transfer learning task because correct response happened in these trials and only correct ones are worth analyzing.

The model is implemented with the PyTorch 1.1 framework on two RTX 2080Ti GPUs. The subject-shared networks consist of 2 ConvLSTM layers, in which the first one has 2 LSTM layers with 8 and 16 hidden units and another one also owns 2 LSTM layers with 16 and 16 hidden units. The subject-specific networks is made of 2 Conv2D layers with 32 and 8 convolution kernels. Note that before entering the subject-specific networks, the output with the size of $ 7 \times 16 \times 32 \times 32$ (not consider the batch-size) is reshaped as $ 112 \times 32 \times 32$. In the end, a Conv2D with 4 kernels and 2 full-connected layers with $4098$ and $512$ hidden units are included in the class prediction networks. Additionally, the learning rate and batch-size are set to 0.0001 and 8, and the optimizer takes Adam, which shows better performance than SGD.

\subsection{Comparison Methods}
We give brief introduction of traditional algorithms and recent deep learning models in this field. For fair comparison, the majority of deep learning-based models share the same setup with the proposed model (Except some have to be set with different configuration architectures). However, considering that the too large feature size may be time-consuming and deteriorate the performance of the traditional algorithms, we make single-frame EEG spectral images (size of $3 \times 32 \times 32$) through observing spectral features from the complete trial duration, and implement an average pooling strategy to reduce the feature size to $3 \times 8 \times 8$.

CNN-3D \cite{R2}: We utilize the convolutional neural network architecture configuration from \cite{R2}, and use the same transfer strategy with CS-DASA. 

TCA \cite{b21}: This work attempts to use the maximum average discrepancy to learn to replicate the transferable components between domains in the kernel Hilbert space. It can reduce the distribution distance between different domains and thus achieve domain adaptation.

W-BDA \cite{b22}: This method adaptively exploits the importance of marginal distribution discrepancy and conditional distribution discrepancy. Meanwhile, it not only considers distribution adaptation, but also adaptively changes the weight of each class.

JDA \cite{b23}: To address the fact that previous transfer methods do not simultaneously reduce the discrepancy between both marginal and conditional distributions, JDA aims to these two kind of distributions in the process of dimensionality reduction, perform domain transfer and establish new feature representations.

DDC \cite{zhang2019cross}: DDC adds an adaptation layer between the source and target domains and sets a domain confusion loss function to allow the network to learn how to classify while reducing the discrepancy in distribution between the source and target domains.

Deep-Coral \cite{b25}: This method applies Coral, an unsupervised domain adaptive method, to deep neural networks in the form of nonlinear transformation that aligns correlations of layer activations. 

Novel-JDA \cite{R16}: Inspired by jointly adapting marginal distributions and conditional distributions, this work proposes a novel JDA method and learn the latent representations through neural networks.

CS-DASA$\triangledown$: To show the improvement resulted from the proposed attention strategy, we also illustrate the performance of the CS-DASA method without the subject-to-subject attention, which is denoted as CS-DASA$\triangledown$.
\subsection{Results and Analysis}
\emph{1) Comparison Results:} We carry out $\mathcal{O}\rightarrow \mathcal{O}$ transfer for all 13 subjects in the dataset, and hence each target subject has another 12 independent source subjects. The statistical results of mean and standard deviation (Mean/STD) are shown in Table \ref{tab1}. For example, the performance of S1 is the mean value of the task set $\left\{\mathbf{S2}\rightarrow \mathbf{S1}, \mathbf{S3}\rightarrow \mathbf{S1},...,\mathbf{S13}\rightarrow \mathbf{S1} \right\}$

In Table \ref{tab1}, SorOnly means the model is only trained on the source domain and is directly tested on the target domain, which belongs to a "nontransfer" baseline.   

For the $\mathcal{O}\rightarrow \mathcal{O}$ transfer scenario, our proposed method improves 16.6\% from the average accuracy perspective compared to SorOnly, which indicates that our proposed method can effectively reduce the domain discrepancy between the source and target domains. Except the SorOnly method, other comparison methods can be categorized as traditional methods and deep learning-based methods. Overall, the traditional methods have only a slight improvement over SorOnly. Even for some target subjects, such as S4, there is a certain degree of decrease in accuracy. This is due to the fact that traditional methods cannot capture the spatial information of EEG images well, and the large dimensionality of the input feature vectors is difficult to be processed by traditional methods. Among a series of traditional methods, W-BDA leads the others by a narrow margin in terms of average accuracy. For the deep learning methods, all the methods showed a significant improvement over the SorOnly and the traditional methods. The DDC method shows its advantage in the $\mathcal{O}\rightarrow \mathcal{O}$ transfer and reaches 60.8\%. Our method beats DDC by 2.8\% and outperforms the other methods except on the S13 subject. In particular, when the target domain is S6, our method leads the second-best method by 2.3\%. After removing the attention mechanism, the performance of the proposed model decreases by 1.1\%, but it still beats the other comparison methods. It is worth noting that the standard deviation of the results is high for each model on this dataset, both for the overall average performance and for the performance on a particular target subject. It indicates that for different source subjects, the transferability varies a lot.

Table \ref{tab2} shows the average accuracy of $\mathcal{M}\rightarrow \mathcal{O}$ transfer with different number of source subjects. Obviously, the performance gap between SorOnly and the deep learning-based transfer method decreases as the number of source subjects increases. When $N_{\mathcal{M}}$ is 9, the transfer learning method improves the accuracy by about 5.3\% at most, which is 8.6\% lower compared to $N_{\mathcal{M}} = 3$. We also note that Novel-JDA beats the other comparison methods when $N_{\mathcal{M}} = 7,9$. Especially when $N_{\mathcal{M}} = 9$, its performance slightly  outperforms CS-DASA$\triangledown$. Overall, our method shows the advantage in each scenario with different source subjects.
\begin{figure*}[htbp]

\centerline{\includegraphics[width=0.95\textwidth]{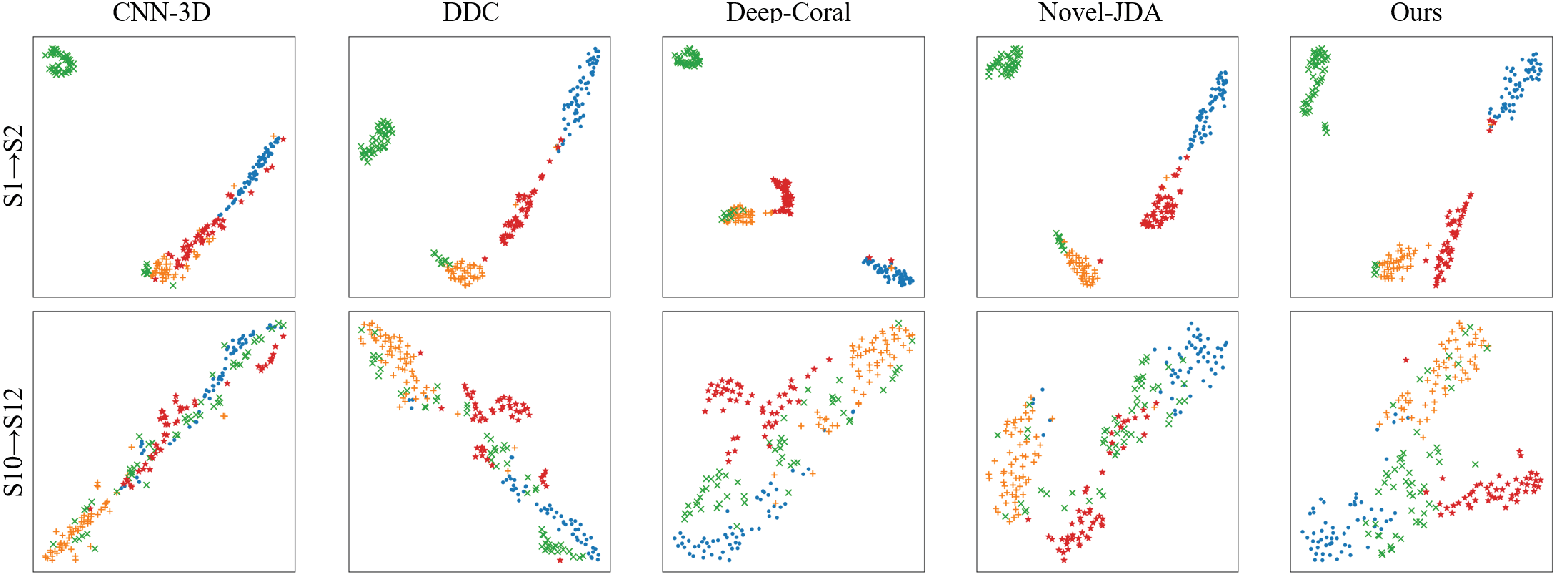}}
\caption{Visualization of features with t-SNE. Blue dots, yellow pluses, green crosses and red stars represent the latent 2-D representations of workload 1-4 after t-SNE. The first row shows the results for the task $S1 \rightarrow S2$ and the second one is the $S10 \rightarrow S12$. Each column demonstrates the results for a transfer learning method.}
\label{tsne}
\end{figure*}

\emph{2) Visualizations:} To intuitively demonstrate the transfer performance of our model, we utilize an effective dimensionality reduction method, t-SNE, to project the features output by the target subjects on 2-D planes. Experiments\cite{b31} show t-SNE can provide better visualization and prevent points gathering in the map center. For simplicity's sake, we select two typical representative tasks with different domain divergence between source and target subjects, where the task $S1 \rightarrow S2$ owns less domain divergence than the task $S10 \rightarrow S12$. Fig. \ref{tsne} shows the t-SNE embedding of these two tasks on each model. We can observe some notable phenomena:
\begin{enumerate}

\item For transfer tasks with less domain divergence between subjects, such as $S1\rightarrow S2$, point clusters for four levels of working memory load are easy to recognize. It indicates that the deep learning-based transfer methods can effectively perform domain adaptation between similar subject features.

\item For both $S1\rightarrow S2$ and $S10\rightarrow S12$ transfer tasks, t-SNE points from CNN-3d show a tendency to mix between different categories and are less discriminative compared to the other four methods. It implies that the model architecture of CNN-3D can not learn transferable features well for domain adaptation compared with the model constructs we built.

\item For the transfer tasks with larger individual differences across subjects, taking $S10\rightarrow S12$ as an example, t-SNE points from CS-DASA can be discriminated better than DDC, Deep-Coral, and Novel-JDA. It shows that the transfer strategy we use makes the samples can be more accurately classified into the correct load classification centers. This phenomenon is also consistent with the fact that CS-DASA can beat other models in terms of accuracy on this task.
\end{enumerate}

\begin{figure}[htbp]

\centerline{\includegraphics[width=0.95\textwidth]{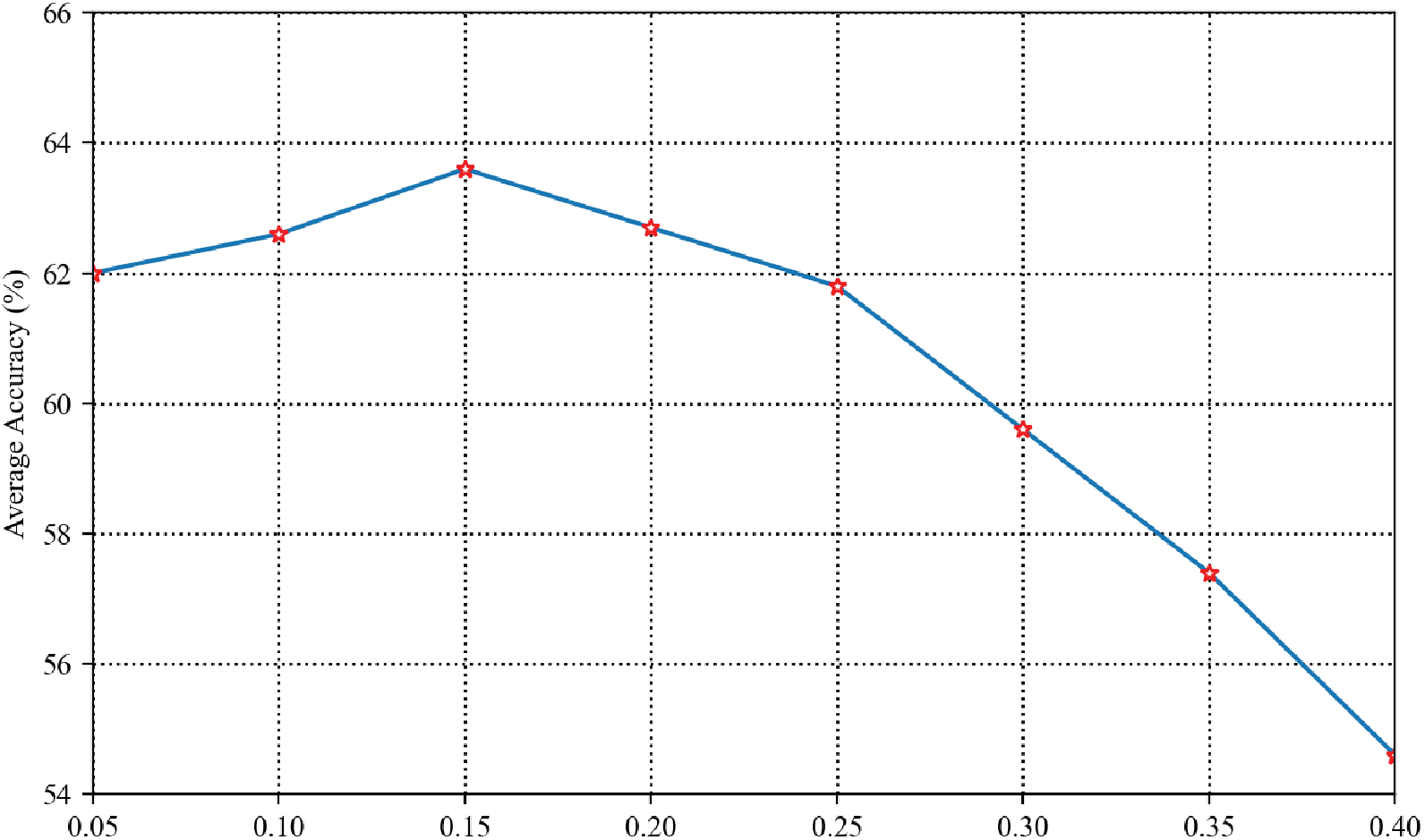}}
\caption{Parameter sensitivity of $\gamma$ }
\label{parameter}
\end{figure}

\emph{3) Parameter Sensitive:} We investigate how the parameter $\gamma$ influence the model performance. Both $\mathcal{O}\rightarrow \mathcal{O}$ and $\mathcal{M}\rightarrow \mathcal{O}$ tasks have been studied, and here we only illustrate the results of $\mathcal{O}\rightarrow \mathcal{O}$ transfer. Fig. \ref{parameter} shows the variation tend of the average transfer classification performance with different $\gamma$ values. It is observed that the model performance in terms of the average accuracy first increases and then decreases as $\gamma$ varies in the ordered set \{0.05,0.10,0.15,0.20,0.25,0.30,0.35,0.40\}. It can be concluded that although a large $\gamma$ value can provide target subjects with more knowledge from source subjects, it doesn't mean the more knowledge is transferred the better performance will be. It is consistent with the fact that only a trade-off between learning features and adapting domain distributions can improve the performance of transfer tasks. 

\begin{figure*}[htbp]
\centerline{\includegraphics[width=0.95\textwidth]{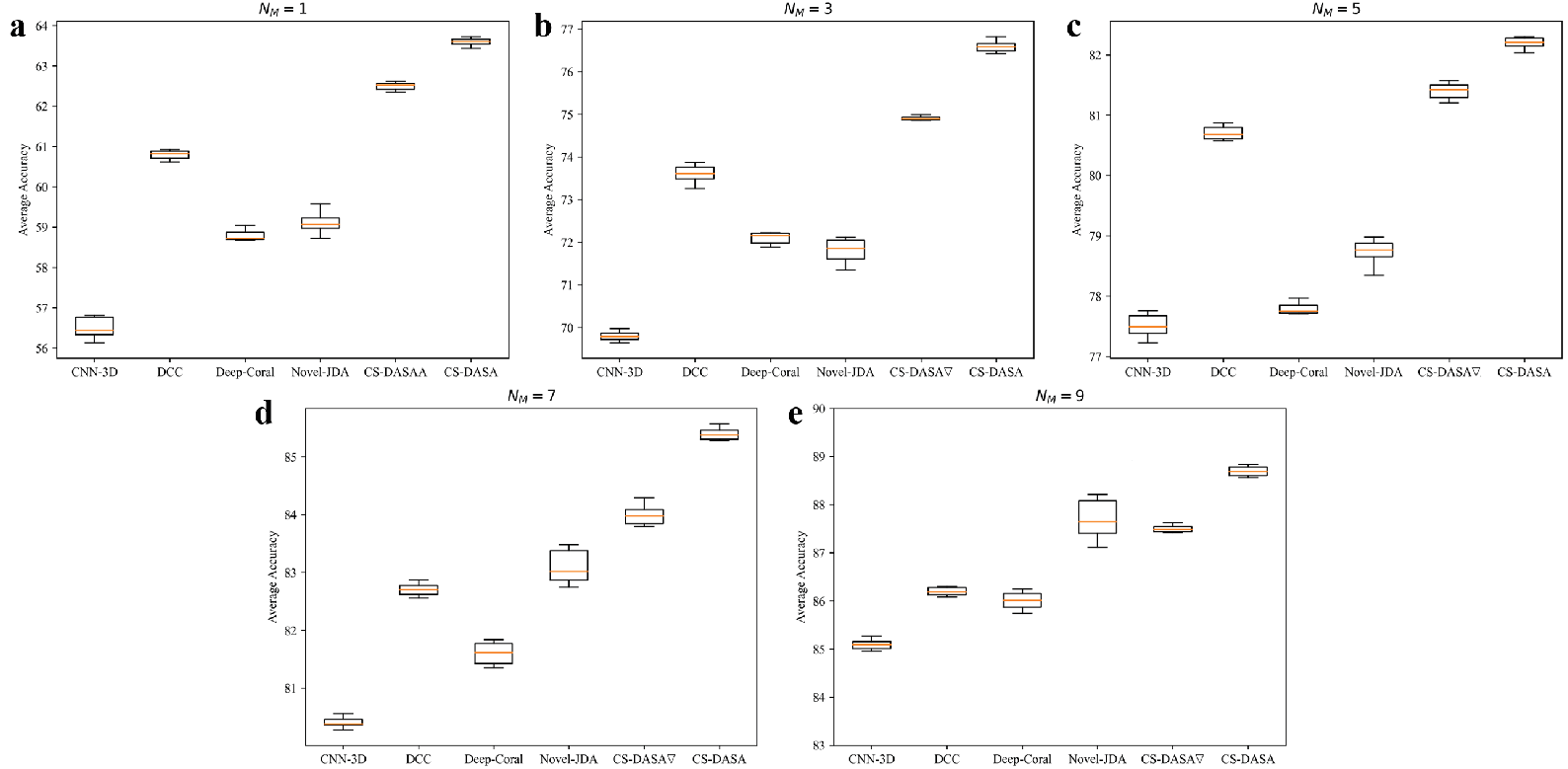}}
\caption{Boxplots for CS-DASA and comparison methods with different $N_{\mathcal{M}}$}
\label{boxplt}
\end{figure*}

\emph{4) Statistical Analysis:} In the cross-subject transfer tasks, the interested evaluation metric is the average accuracy for all the sub-tasks since it can represent the overall performance of a transfer method and it is hard to evaluate a huge number of sub-tasks. To further demonstrate the performance in terms of statistics, we illustrate the average accuracy of each independent run with different settings of $N_{\mathcal{M}}$ through boxplots shown in Fig. \ref{boxplt}. Obviously, compared with other methods, Novel-JDA shows the more unstable results in all the settings of $N_{\mathcal{M}}$, which might be introduced by its adversarial training strategy.

Also, we perform the Wilcoxon test \cite{demvsar2006statistical} to discriminate the average accuracy difference in the view of the statistical significance. The results show that under all the settings of $N_{\mathcal{M}}$, CS-DASA, the proposed method, beats all the other comparison methods significantly ($p<0.05$). Besides, when $N_{\mathcal{M}} = 9$, the results indicate that the CS-DASA$\triangledown$ method cannot significantly outperform the Novel-JDA method ($p=0.169$), while CS-DASA$\triangledown$ shows better performance than Novel-JDA ($p<0.05$) under other $N_{\mathcal{M}}$ settings. It can be concluded that these two methods might share similar performance when the number of source subjects is large enough and Novel-JDA can achieve a not bad performance with a wider source subject sample distribution.

\section{Discussion}
In EEG-based workload classification tasks, building supervised learning models is a common solution. However, in practical application scenarios, it is difficult or costly to obtain enough labeled data to calibrate. Besides, since the feature difference between subjects may be large, the EEG signals from the subjects out of the existing database are easy to fail in the trained model. To solve the above problems, we design a novel transfer learning classification method and conduct a lot of experiments to verify the effectiveness of the proposed method. The experimental results are quite inspiring, and we make some conclusions.

\begin{enumerate}

\item In the $\mathcal{O}\rightarrow \mathcal{O}$ transfer scenario, traditional transfer learning models lag behind deep learning-based methods by about 20\% and only slightly outperform non-transfer method SorOnly. These traditional methods have been reported to achieve good results on certain data sets, but fail on this WM classification task in this work. The main reason is that these methods rely heavily on feature design (including channel selection, feature dimensionality reduction, etc.). Therefore, it is worthwhile to apply these methods when the transfer task does not require an end-to-end structure and the researcher has the experience to design the features.

\item In general,  the average performance of the $\mathcal{M}\rightarrow \mathcal{O}$  transfer is better than that of the $\mathcal{O}\rightarrow \mathcal{O}$  transfer.  Specifically, with the value of $N_\mathcal{M}$ increasing, the performance of our model becomes better. It can be explained that when the source domain includes more samples from different subjects, the labeled EEG data from various distributions is abundant, and the model can learn more transferable features.

\item In both $\mathcal{O}\rightarrow \mathcal{O}$ and $\mathcal{M}\rightarrow \mathcal{O}$ transfer scenarios, the standard deviation of the accuracy of the various methods is large (greater than 20\%). In addition, there is a large difference in the performance of the models when the target subjects are different. For example, in one-to-one transfer, the gap in the average accuracy between the target subjects S4 and S5 using the CS-DASA method is about 40\% compared to S13. It indicates that there are subject groups with similar feature distribution as well as subject groups with very large feature space divergence among these 13 subjects. Therefore, it is important to select the appropriate source subjects to improve the model performance.

\item The Novel-JDA doesn't show a satisfactory result in the $\mathcal{O}\rightarrow \mathcal{O}$ transfer task, and the average accuracy is 4.5\% lower than CS-DASA. However, with the number of source subjects increasing, the gap drops down to 1.0\% when there exist 9 subjects in the source domain. One reasonable explanation is that the amount of samples in each subject is small, and the adversarial training strategy might fail when the distribution diversity of the source EEG data is not abundant.  
\end{enumerate} 

Concerning the problems encountered in this article and the current development in this field, two possible future research directions can be explored:

\begin{enumerate}

\item \emph{Automatic selection of the source subjects:} When implementing cross-subject transfer tasks, the selection of the source subjects is quite important. A possible solution is to design a cross-subject feature evaluation algorithm. Based on the evaluation scores, a subset of alternative subjects is selected to serve as the source of the current target subject. This strategy can avoid using too many labeled samples and can improve the accuracy of the classification task to some extent.

\item \emph{Subdomain adaptation network for cross-subject WM load classification:}  The transfer methods mainly used in the EEG-based WM tasks focus on the global domain shift. And hence, the fine-grained information for each workload level is neglected, which might lead to the confusion of different categories within each domain.

\end{enumerate}

\section{Conclusion}

This work targets the EEG working memory load classification task and presents a novel domain adaptation method based on deep learning. The Azimuthal Equidisatant projection is utilized to convert the 3-D EEG electrode positions into 2-D positions, which can preserve more spatial features containing topology information. Additionally, we apply a global spatial attention mechanism to ensuring the model better understand features from the target domain. Comprehensive experiments including some traditional and recent transfer learning methods show the proposed method can achieve superior performance in terms of the accuracy.
It is also revealed that the spatial attention mechanism can effectively improve the overall performance. Transfer learning for EEG WM load tasks is rarely explored, and our work can encourage and promote the development and the practical applications of this topic.

\backmatter

\section*{Declarations}
\begin{itemize}
\item \textbf{Funding} This work was supported by Postgraduate Research and Practice Innovation Program of Jiangsu Province under Grant KYCX24\_0609 and Science and Technology Innovation 2030-Key Project of ``New Generation Artificial Intelligence'' under Grant 2021ZD0113103.
\item \textbf{Conflict of interest}
The authors declare that there are no relevant financial or non-financial interests that could be construed as a potential conflict of interest with respect to the work described in this manuscript.

\item \textbf{Data availability} The datasets used in this research are publicly available, and datasets can be made available by contacting the corresponding author.

\item \textbf{Author contribution} Junfu Chen played a pivotal role in the conceptualization and execution of the experiments, as well as in drafting the manuscript. Sirui Li reproduced comparison experiments and helped with writing. Dechang Pi contributed significantly by providing a thorough review and making critical revisions to the manuscript. All authors have collaboratively worked on the article, and have given their approval for the final version submitted for publication.
\end{itemize}


\bibliography{main}

\end{document}